\documentclass[letterpaper,twocolumn,10pt]{article}
\usepackage{usenix-2020-09}

\usepackage{tikz}
\usepackage{amsmath}
\usepackage{url} 
\usepackage{hyperref}
\makeatletter
\def\UrlAlphabet{%
      \do\a\do\b\do\c\do\d\do\e\do\f\do\g\do\h\do\i\do\j%
      \do\k\do\l\do\m\do\n\do\o\do\p\do\q\do\r\do\s\do\t%
      \do\u\do\v\do\w\do\x\do\y\do\z\do\A\do\B\do\C\do\D%
      \do\E\do\F\do\G\do\H\do\I\do\J\do\K\do\L\do\M\do\N%
      \do\O\do\P\do\Q\do\R\do\S\do\T\do\U\do\V\do\W\do\X%
      \do\Y\do\Z}
\def\UrlDigits{\do\1\do\2\do\3\do\4\do\5\do\6\do\7\do\8\do\9\do\0}
\g@addto@macro{\UrlBreaks}{\UrlOrds}
\g@addto@macro{\UrlBreaks}{\UrlAlphabet}
\g@addto@macro{\UrlBreaks}{\UrlDigits}
\makeatother

\usepackage{array}    
\usepackage{threeparttable} 
\usepackage{subcaption} 
\usepackage{booktabs} 
\usepackage{tabularx} 
\usepackage[linesnumbered,ruled,vlined]{algorithm2e} 
\usepackage{comment} 
\usepackage{multirow} 
\usepackage{filecontents}

\begin{document}

\date{}


\title{\Large \bf EPS-MoE: Expert Pipeline Scheduler for Cost-Efficient MoE Inference}

\author{
    {Yulei Qian, Fengcun Li, Xiangyang Ji, Xiaoyu Zhao, Jianchao Tan, Kefeng Zhang, Xunliang Cai}\\
    {
        \{qianyulei02, lifengcun, jixiangyang, zhaoxiaoyu17, tanjianchao02, zhangkefeng, caixunliang\}@meituan.com
    }
}

\maketitle

\begin{abstract}

The Mixture-of-Experts (MoE) model has emerged as a prominent architecture in the field of Large Language Models (LLMs), providing a better balance between model performance and computational efficiency. However the General Matrix Multiply (GEMM) operations and large parameters introduce challenges related to computational efficiency and communication overhead, which become throughput bottlenecks during inference. Applying a single parallelism strategy like EP, DP, TP or a straightforward combination of them to MoE usually achieves sub-optimal inference throughput.
This paper introduces EPS-MoE, a novel expert pipeline scheduler for MoE that surpasses the existing parallelism schemes. Our approach optimizes the computation of MoE FeedForward Network (FFN) modules by dynamically selecting the best kernel implementation of GroupGemm and DenseGemm for different loads and adaptively overlapping these computations with communication, leading to a substantial increase in throughput. Our experimental results demonstrate at most 52.4\% improvement in prefill throughput compared to existing parallel inference methods. Specifically, our method accelerated the highly optimized DeepSeekV2 model from a claimed 100K tokens per second to at least 120K tokens per second.

\end{abstract}

\section{Introduction}

\begin{table}[htbp]
\caption{Prefill Throughput Gains of EPS-MoE.}
\label{tab:prefill_benefit_overall}
\centering
    \begin{threeparttable}
    \begin{tabular}{l|l}
        \toprule
        Model & Prefill Throughput \\
        \midrule
        DeepSeekV2 & 100 k (token/s) \\
        DeepSeekV2 (EPS-MoE) & 121.8 k (token/s) , +21.8\%\\
        Mixtral (vllm) & 71.84 k (token/s) \\
        Mixtral (vllm + EPS-MoE) & 94.89 k (token/s), +32.2\%\\
        DBRX(vllm) & 37.23 k (token/s) \\
        DBRX(vllm + EPS-MoE) & 56.74 k (token/s), +52.4\%\\
        \bottomrule
    \end{tabular}
    \begin{tablenotes}
        \item[1] DeepSeekV2 and DBRX were tested on 8xH800-80GB SXM, Mixtral(8x7B) tested on 4xH800-80GB SXM. 
    \end{tablenotes}
    \end{threeparttable}

\end{table}

The remarkable capabilities of the LLM have attracted various organizations to devote resources to optimize their architectures for better performance and efficiency, leading to the development of advanced MoE architectures like Mixtral\cite{mixtral}, DBRX\cite{dbrx}, DeepSeekV2 \cite{deepseekv2}, Grok\cite{grok}, Gemini 1.5\cite{gemini1.5}, Snowflake\cite{snowflake_arctic}\cite{snowflake_arctic_paper} and others\cite{xverse_moe_a36b}\cite{hmoe}\cite{stmoe2022}\cite{fastmoe2021}\cite{flexmoe2023}\cite{hybridmoe2023}\cite{scatteredmoe2024}\cite{deepspeedmoe2022}\cite{hetumoe2022}. These architectures offer significant advantages by enabling the dynamic selection of specialized experts, thus optimizing performance and computational efficiency. MoE models have demonstrated the ability to scale up model parameters significantly for improved performance while maintaining a manageable computational footprint.
Typically, MoE incorporates a gating mechanism that directs the output of the attention mechanism to a subset of experts, thereby activating only a fraction of the model's parameters. This approach can expand model capacity with far fewer activated parameters to achieve performance comparable to larger dense models. For instance, DeepSeekV2, with only 21 billion activated parameters, rivals the performance of Llama3's 70 billion parameters. \cite{deepseekv2}

However, MoE architectures encounter significant challenges when scaling to accommodate large sequence lengths and batch sizes. For example, Mixtral 8x7B requires only 12.6 billion activated parameters per token but can demand up to 46 billion parameters for large batch sizes.  Due to the large total number of parameters, MoE models often require multi-GPU parallel inference, which also leads to an increase in communication time.
Moreover, the router's top-k gating mechanism, while beneficial for selecting relevant experts, intensifies the communication when k is large. In such scenarios, the communication requirement can be magnified k-fold, as information must be exchanged with k different experts simultaneously. This can result in bottlenecks in the inference pipeline, as the computation for each expert's output cannot commence until the communication of inputs is complete. 
In addition to communication challenges, MoE models also face issues with low computation density, particularly when the distribution of tokens leads to fragmented workloads across the experts. The resulting imbalance can leave some computational resources underutilized, further impacting the overall efficiency of the inference process.

\begin{figure}
    \centering
    \includegraphics[width=0.95\linewidth]{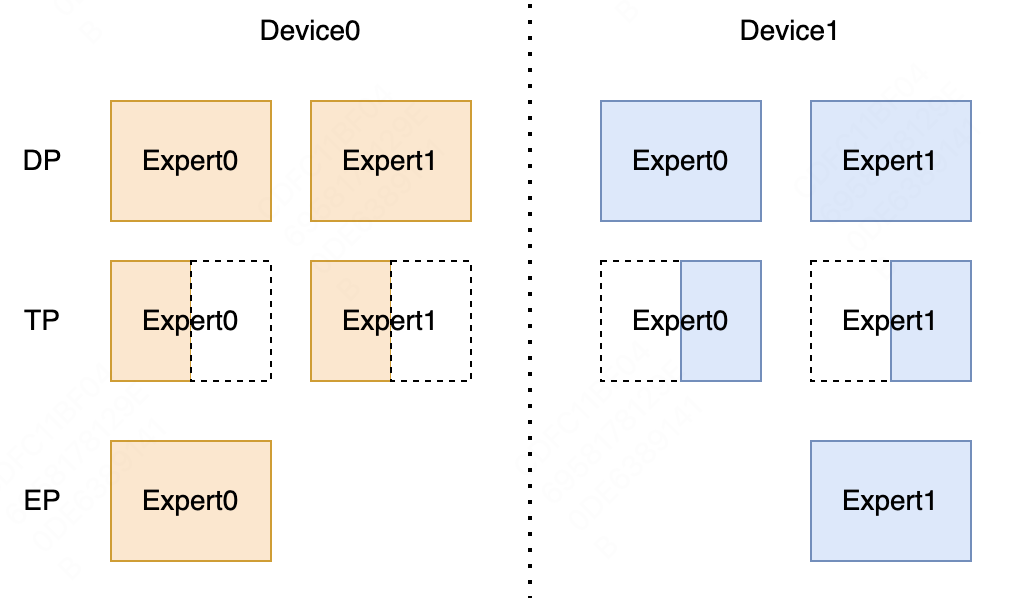}
    \caption{Weight partition of DP, TP and EP for two devices and two experts.}
    \label{fig:dp_tp_ep_pp}
\end{figure}

Due to the rapid scaling of the model parameters, distributed serving architectures have become indispensable for serving MoE models at scale. Common strategies \cite{megatron-lm}\cite{nvidia_parallelisms} include Data Parallelism (DP), Tensor Parallelism (TP) and Expert Parallelism (EP), as shown in Figure~\ref{fig:dp_tp_ep_pp}. Each method targets different aspects of model serving, such as reducing communication overhead and enhancing computational efficiency. However, a single strategy or a straightforward combination of them cannot obtain optimal inference throughput.

To address these inference challenges from MoE architectures and go beyond these suboptimal solutions, we propose EPS-MoE, a novel expert pipeline scheduler for efficiently serving MoE architectures. The framework consists of three main highlights. 
\textit{1) Parallel Strategy} With a theoretical analysis, we choose to apply DP or TP on Attention blocks based on the Attention algorithms and EP on MoE blocks. We provide a detailed analysis in the following section.  \textit{2) Expert Pipeline Scheduler} We demonstrate a new tensor or weight split method to achieve better memory I/O performance and to take the advantage of switching from \textit{GroupGemm} to \textit{DenseGemm}\footnote{For convenience, we use \textit{GroupGemm} to refer to grouped GEMM from \textit{cutlass} and \textit{DenseGemm} to refer to \textit{cublas} GEMM for dense matrix multiplication.}. Based on this split method, we propose the expert pipeline scheduler to submit experts sequentially to a kernel for computation. We overlap this sequential computation with pipeline parallel to address the overhead. \textit{3) Computation and Communication Overlapping} With a pipeline between computation and communication at the kernel level, we will show how EPS-MoE achieves better performance on the inference of MoE models. 
We summarize the core contributions as follows:
\begin{itemize}
\item \textbf{Contribution 1:} We introduced a novel expert pipeline parallel scheduler for efficient MoE model inference, which involves a fine-grained overlapping between computation and communication at the kernel level. 
\item \textbf{Contribution 2:} We conducted an in-depth analysis of GEMM efficiency and designed a horizontal split for inputs and an expert split for MoE weights. Based on these splitting methods, we implemented a switching strategy from \textit{GroupGemm} to \textit{DenseGemm} under specified loads to improve computational efficiency.
\item \textbf{Contribution 3:} We explored different concurrency modes for Attention and MoE: TP+TP, DP+EP, and TP+EP and analyzed the efficiency of these modes.
\item \textbf{Contribution 4:} We applied our method to a variety of different models for testing. The experiments showed that our method can improve the prefill throughput 52.4\% at most.
\end{itemize}

\begin{figure*}[htbp]
    \centering
    \includegraphics[width=1\linewidth]{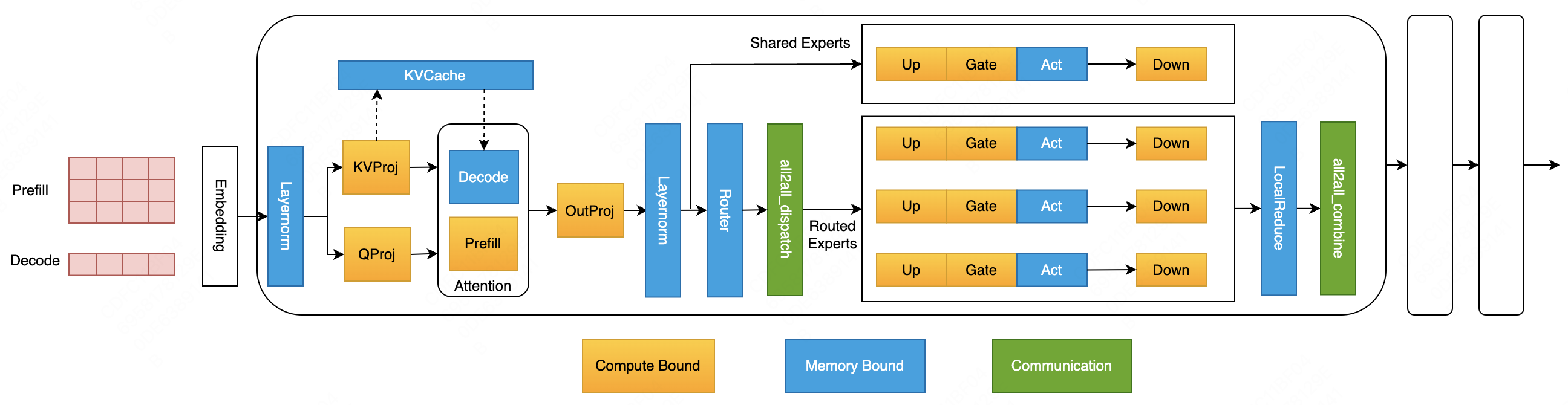}
    \caption{MoE architecture. Adpoted from\cite{nanoflow}. The operations in the yellow boxes are compute-bound, mostly \textit{GEMMs}. The light blue box operations are memory-bound. The operations in the green boxes are communication operations.}
    \label{fig:transformer_architecture}
\end{figure*}

\section{Related Work}
\subsection{MoE architectures}
The groundbreaking work by Shazeer et al.~\cite{shazeer2017outrageously} introduced the Sparsely-Gated Mixture-of-Experts (MoE) layer, which laid the foundation for scaling neural networks by utilizing a sparse activation pattern. This approach allows for efficient training of large models with a mixture of experts, where each expert is only activated for specific inputs. However, the original MoE layer suffers from challenges in balancing the load among experts and may lead to the underutilization of some experts. 
Fedus et al.\cite{gshard} proposed GShard, a method to scale giant models by employing conditional computation and automatic sharding. GShard addresses some of the limitations of the original MoE by enabling dynamic routing and sharding of experts across different devices, thus improving scalability. Nonetheless, GShard may face difficulties in maintaining model coherence across shards and requires sophisticated infrastructure to manage the distributed computation.
Switch Transformers by Fedus et al.~\cite{fedus2021switch} takes sparsity to the next level by introducing a simple and efficient sparsity pattern that allows scaling to trillion-parameter models. The Switch Transformers utilizes a gating mechanism to activate experts based on the input data, which can significantly reduce computational overhead. However, the gating mechanism adds complexity to the model, and the benefits of sparsity may diminish as the model size increases.
Besides the advanced MoE foundational models such as Mixtral~\cite{mixtral}, DB
RX~\cite{dbrx}, DeepSeekV2~\cite{deepseekv2}, Grok~\cite{grok}, Gemini 1.5~\cite{gemini1.5}, there are some novel parallel designs on MoE architectures. 
ScMoE~\cite{scmoe} and Snowflake Arctic~\cite{snowflake_arctic}~\cite{snowflake_arctic_paper} proposed to add a long shortcut for parallelism between the multi-experts branch and the whole dense branch. These designs significantly enhance both training and inference speeds. Nevertheless, the reliance on these specific network topologies for shortcuts might cause sub-optimal inference efficiency in some scenarios.

\subsection{Disaggregated PD Serving}

 Disaggregated PD serving technology is a popular architecture recently, such as DistServe\cite{dist_serve}, PDServe\cite{pd_serve}, SplitWise\cite{split_wise}, MoonCake\cite{mooncake}, and others\cite{disaggregate_llm}. The tasks in the prefill and decode stages of LLM inference are different: prefill is a compute-bound task constrained by the TTFT \footnote{TTFT: Time To First Token} SLO, while decode is memory-bound constrained by the TPOT \footnote{TPOT: Time Per Output Token} SLO. If these two stages are disaggregated to run on different serving instances, each can be optimized based on its respective SLO, thereby fully utilizing machine resources. 
However, the disaggregated architecture cannot solve the issue of kernels' execution efficiency in MoE inference, nor can it address the problem of high communication overhead. As shown in Figure~\ref{fig:transformer_architecture}, whether in the prefill or decode stage, there are many memory-bound and communicating kernels in their respective execution processes. Regardless of the increment in the computational density of a single stage, these memory-bound kernels cannot be completely eliminated, and therefore, the computational resource utilization of that stage cannot be fully improved. Additionally, both the prefill and decode stages face the overhead brought by communication. The more layers the model has or the more GPUs the model uses, the higher the proportion of communication time is.

\subsection{Operator-level Parallelism}
Operator-level parallelism is a solution that takes advantage of the load characteristics of operators to achieve parallelism. Nanoflow\cite{nanoflow} implemented pipeline parallelism between operators by splitting the input into finer-grained batches. Nanoflow correctly pointed out that the computational efficiency of operators and the number of SMs are not always linearly related; as computational efficiency increases, adding more SMs does not significantly improve computational efficiency. ScMoE \cite{scmoe} additionally implemented kernel-level overlapping strategies for MoE to reduce communication-related overhead. However, these optimization methods do not design the optimal pipeline scheduling strategy based on the GEMM (General Matrix Multiply) characteristics of MoE, nor do they design parallel solutions that leverage the model characteristics of MoE. As we will see in the subsequent analysis, if operator-level parallelism does not fully utilize the characteristics of MoE, it can lead to repeated memory I/O of parameters. Furthermore, MoE inference is highly sensitive to load, ignoring load characteristics and implementing fixed scheduling strategies often fails to achieve the best overall results.

\section{Analysis}
The inference latency of LLMs is influenced by various factors, and numerous studies have attempted to model this ~\cite{sarathi} \cite{llm_inference_unveiled} \cite{vidur}. The traditional optimization approach maximizes the computational efficiency of each kernel within its respective types. For instance, if a kernel is compute-bound or memory-bound under a certain workload, ideally, it should fully utilize the hardware's computational FLOPs or memory bandwidth. Achieving such an ideal scenario is extremely challenging. As we will see in the subsequent analysis, even the disaggregated architectures that separate workloads into prefill and decode stages cannot completely resolve the issue of balanced efficiency at the kernel level. This is because both the prefill and decode stages are composed of kernels from multiple types of workloads as shown in Figure~\ref{fig:transformer_architecture}, and this compositional relationship does not change with the disaggregation of prefill and decode. Next, we will analyze the inference issues of MoE models from the perspectives of three types of kernels: computation, memory access, and communication.

\begin{figure}[htbp]
    \centering
    \includegraphics[width=1\linewidth]{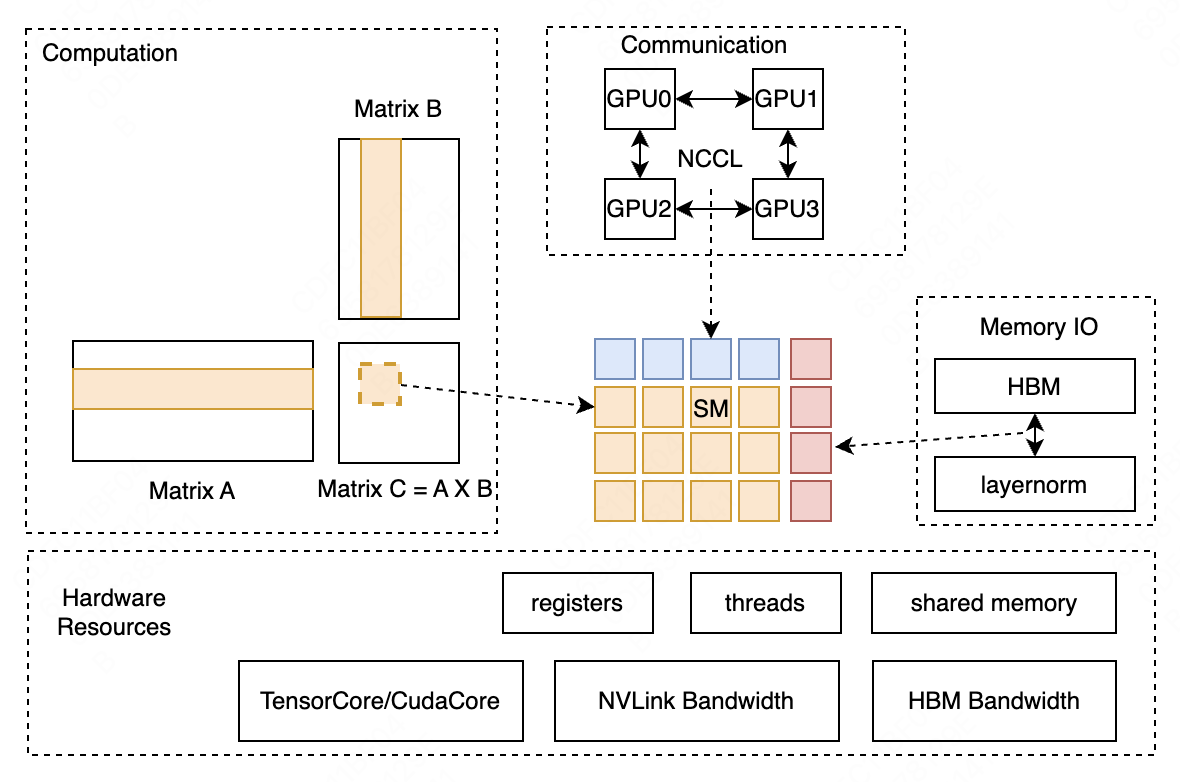}
    \caption{Resources view of Nvidia GPU Architecture}
    \label{fig:gpu-arch}
\end{figure}

\subsection{Modern GPU Architecture}

We use NVIDIA A100-80GB SXM and H800-80GB SXM GPUs to do our tests. The matrix computation libraries used are mainly \textbf{cutlass} \cite{cutlass} and \textbf{cublas} \cite{cublas}. NVIDIA's matrix computations are primarily conducted using a tiling strategy. The input matrix and weight matrix are divided into different tiles along rows or columns. Each block processes one or more tiles, and each block runs on a single SM (Streaming Multiprocessor) as shown in Figure~\ref{fig:gpu-arch}. The SM forms the basic computational unit. All hardware resources like Tensor Cores or CUDA Cores, memory bandwidth, and communication bandwidth are accessed through \textit{SMs}.

\subsection{Computation}
\begin{figure}
    \centering
    \includegraphics[width=1\linewidth]{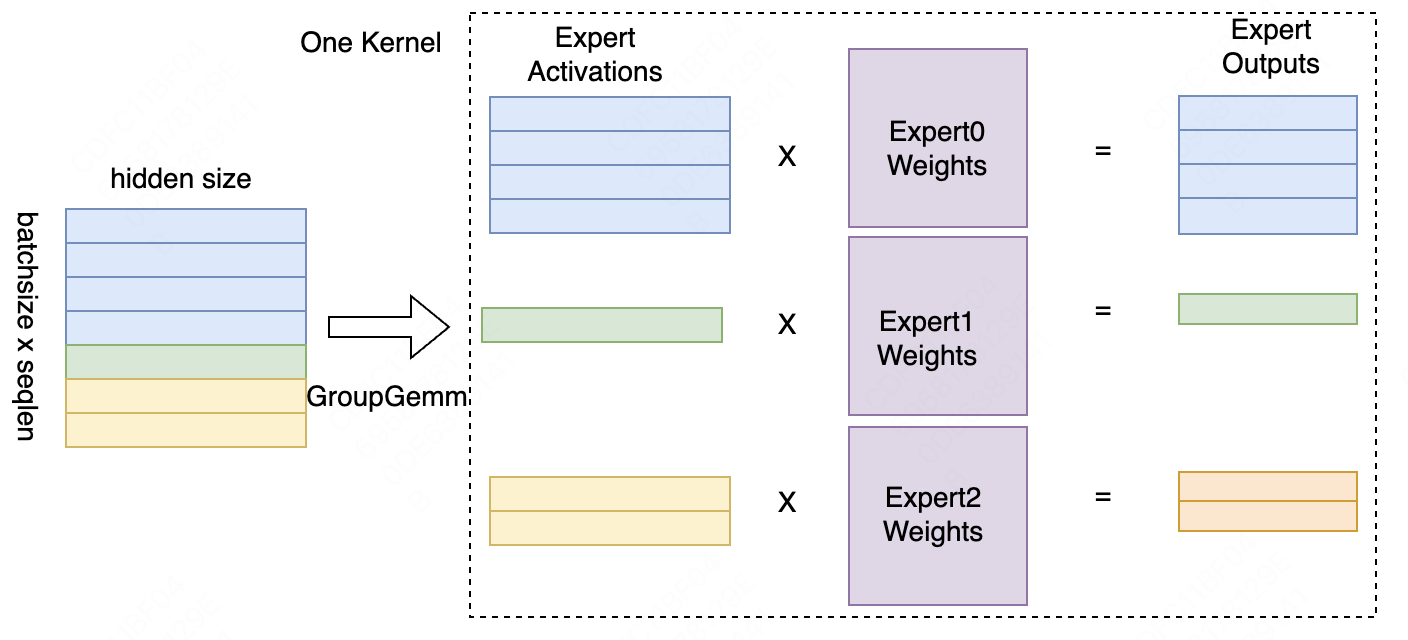}
\caption{\textit{GroupGemm} demonstration. Adpoted from \cite{who_says_elephants}. All matrix multiplication operations are performed through a single kernel launch.}
    \label{fig:groupgemm_demo}
\end{figure}

\begin{figure*}[htbp]
\centerline{\includegraphics[width=1\linewidth]{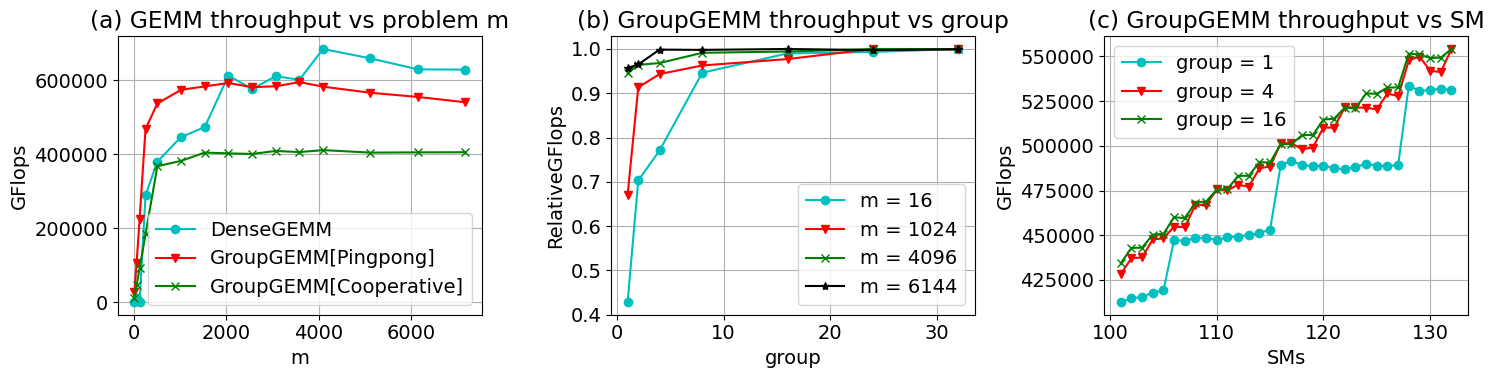}}
\caption{GEMM profiling data. (a) Throughput of different GEMMs was tested with varying input sizes, focusing on \textit{Gate} and \textit{Up} matrices from MoE blocks. For \textit{GroupGemm}, 16 Experts were used with matrix dimensions $[1536, 5120]$, ensuring equal total computation with \textit{DenseGemm}. (b) \textit{GroupGemm [Pingpong]} was tested for relative throughput across different groups and problem sizes, comparing current group throughput to the maximum. (c) \textit{GroupGemm [Pingpong]} was tested with input size $m = 6144$, evaluating throughput across different SM counts and groups.}

\label{fig:mfu_of_gemm}
\end{figure*}

The inference performance of the MoE model is significantly affected by the load scenario. Let \( E \) be the number of experts in the MoE model, and \( k \) be the number of experts selected by each token in a single forward pass. When the number of tokens in a single forward pass is \( m \), the number of activated experts in the MoE model satisfies the following relationship:
\[ExpectedActivatedExperts = (1-(1-\frac{k}{E})^m) \times E\]

When \( m \) is relatively large, such as in the prefill stage, all experts in the MoE are activated. At this time, the MoE model inference is compute-bound. It can fully take advantage of having relatively small per-token FLOPs, thereby accelerating inference in the prefill stage.

When \( m \) is relatively small, such as in the decode stage, the MoE inference becomes memory-bound. As \( m \) increases within a certain range, the number of activated experts in the model also increases, leading to a higher amount of weight parameters that need to be loaded. 

Therefore, the inference of a MoE model needs to consider the load conditions of its application scenarios.

MoE typically uses \textit{GroupGemm} for computation, as shown in Figure~\ref{fig:groupgemm_demo}~\cite{who_says_elephants}. Generally speaking, \textit{GroupGemm} is a relatively efficient kernel. However, we have found that as the computation load increases, the performance of the \textit{GroupGemm} kernel does not remain constant. We conducted some tests on \textit{GroupGemm} and \textit{DenseGemm} based on real model settings, as shown in Figure~\ref{fig:mfu_of_gemm}, and derived several important conclusions.


\textbf{Conclusion 1: The computational efficiency of \textit{GroupGemm} and \textit{DenseGemm} varies with input size, each having advantages in different ranges.}

As shown in Figure~\ref{fig:mfu_of_gemm}(a), the computation efficiency of \textit{GroupGemm} and \textit{DenseGemm} gradually improves as the input \( m \) increases. Furthermore, when \( m \ge 4096\), the computation efficiency of \textit{GroupGemm} will be lower than that of \textit{DenseGemm} with the same computation load. When \( m < 2048 \) , the efficiency of \textit{GroupGemm} is higher than that of \textit{DenseGemm}. This indicates that \textbf{the optimal kernel implementation changes with the input scale}. This is known as kernel tuning for different problem shapes. However, we brought GEMM type into consideration in this paper. In the inference process of MoE models, variations in input load are common. For example, during the prefill stage, the input often falls within the range where \textit{DenseGemm} is faster; whereas, during the decode stage, the input usually falls within the range where \textit{GroupGemm} is faster.

\textbf{Conclusion 2: For \textit{GroupGemm}, once the input size reaches a certain number, having more groups does not lead to higher throughput.}

Increasing the number of groups in \textit{GroupGemm} has always been a method to improve computational throughput, especially in the fine-grained MoE model. This approach always works when the input size is relatively small. However, when the input size is relatively large, having more groups does not further increase computational throughput, as shown in Figure~\ref{fig:mfu_of_gemm}(b). Therefore, when the input size is relatively large, splitting a larger \textit{GroupGemm} into some smaller \textit{GroupGemm}s does not lead to a decrease in throughput.

\textbf{Conclusion 3: For \textit{GroupGemm}, once the input size reaches a certain number, using more SMs does not lead to higher throughput.}

Nanoflow\cite{nanoflow} has pointed out that for \textit{DenseGemm}, the number of SMs it occupies can be reduced within a certain range without affecting GEMM's computational throughput. For \textit{GroupGemm}, we can observe a similar trend from Figure~\ref{fig:mfu_of_gemm}(c). Additionally, the tested results also show that for \textit{GroupGemm}s with different numbers of groups  reducing the number of SMs occupied by GEMM within a certain range also does not affect GEMM's execution efficiency.

\begin{figure}[htbp]
    \centering
    \includegraphics[width=1\linewidth]{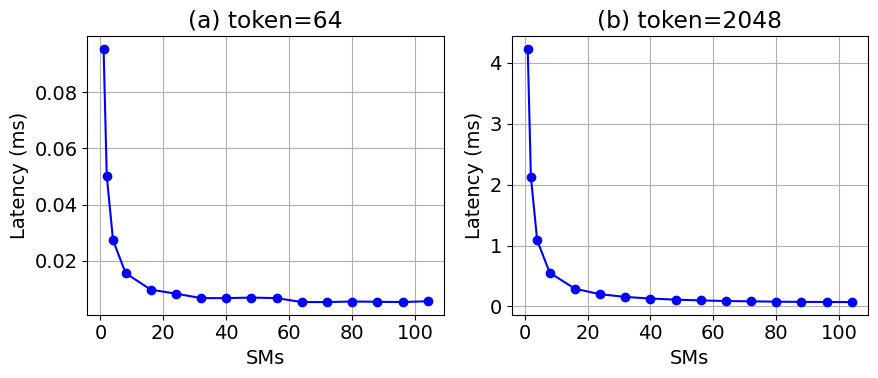}
    \caption{Latency of the memory-bound \textit{silu\_activation} kernel with varying loads and SMs: (a) Tensor [64, 8192] - performance degradation is negligible beyond 40 SMs. (b) Tensor [2048, 8192] - performance degradation is negligible beyond 60 SMs.}
    \label{fig:io_bound_kernel_sm}
\end{figure}

\subsection{Memory Access}

During MoE inference, there are many memory-bound operators, such as \textit{layernorm}, \textit{residual}, \textit{activation}, \textit{topKGating}, etc and there are many traditional optimization methods such as kernel fusion, weight quantization and vectorized memory access ~\cite{cuda_pro_tip}. Another memory-bound operator is Attention KVCache loading, many papers have studied this issue.
In principle, memory-bound kernels also require more SMs to achieve optimal performance. However, practical tests on the \textit{silu\_activation} operator show that while the execution time decreases with an increase in SMs, the performance gains become negligible for MoE inference once the SM count exceeds a certain threshold. As illustrated in Figure~\ref{fig:io_bound_kernel_sm}, this threshold is closely related to the operator's computational load. When the number of tokens is 64, 40 SMs are sufficient to achieve good kernel performance. When the number of tokens is 128, 256, or 2048, 60 SMs perform adequately.
When the input rows is small, MoE GEMM will result in memory-bound. We can find a similar trend in such situation.

\subsection{Communication}

\begin{figure}
    \centering
    \includegraphics[width=1\linewidth]{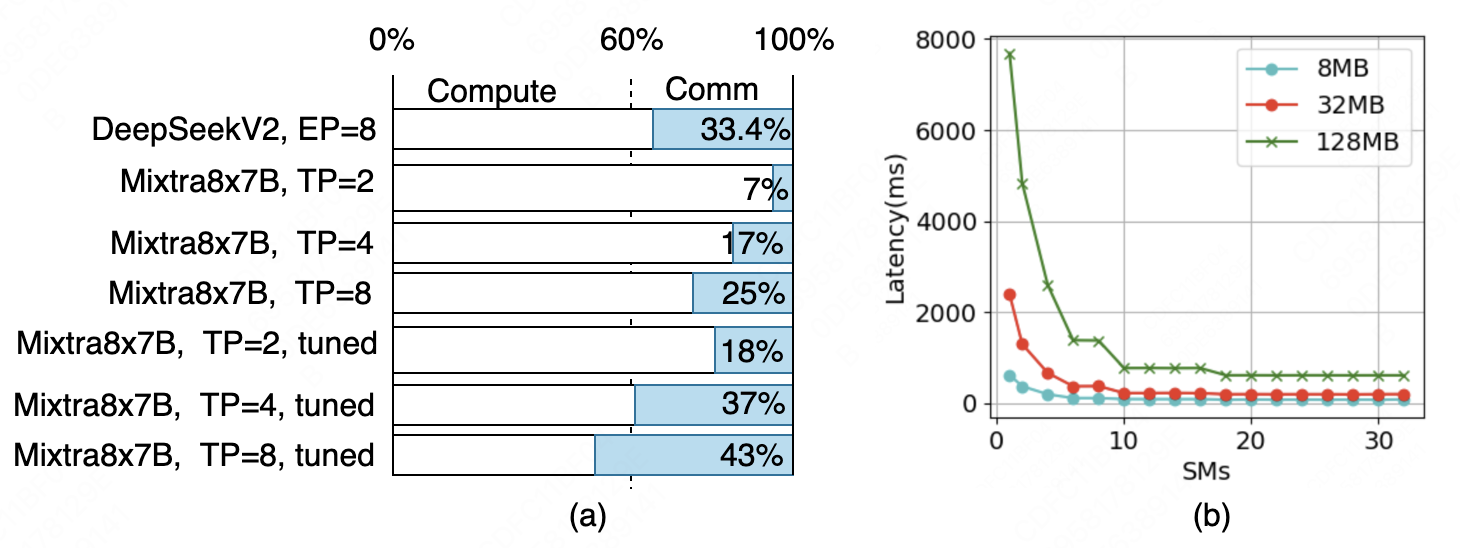}
    \caption{(a) Time Consumption of Communication and Computation across Models. (b) Latency of \textit{all2all} kernel with varying loads and SMs. 10-20 SMs are sufficient, with no significant latency improvement beyond that. Tested on NVIDIA A100-80GB SXM with NVLink.}
    \label{fig:time_consumption}
\end{figure}

Communication is a common issue in LLM inference because the model size is often too large to be deployed by a single GPU. However, multi-GPU parallel computation also introduces additional communication overhead between the GPUs.
From Figure~\ref{fig:time_consumption}, it can be seen that the proportion of time spent on communication increases with more GPUs. If we tune the kernel to make \textit{GroupGemm} more efficient, the proportion of communication time will become much more significant. To reduce communication time, we can use FP8 quant to decrease the communication volume, striking a balance between precision and latency. However, this does not completely solve the problem; as batch size increases, the communication volume will further increase. Moreover, as the model size grows, multi-machine parallel strategies will be introduced, where the bandwidth between machines will be much lower than that between GPUs in a single machine. At this point, the proportion of time spent on communication will become even more significant.
Communication consumes SM resources, but we have observed that within a certain range, adjusting the number of SMs allocated to communication kernels has almost no impact on communication throughput, even in  scenarios with high communication volumes such as in prefill, as shown in Figure~\ref{fig:time_consumption}.

\section{Design of EPS-MoE}

\subsection{Overview of EPS-MoE}

\begin{figure*}[htbp]
    \centering
    \begin{subfigure}[b]{0.45\textwidth}
        \centering
        \includegraphics[width=1\linewidth]{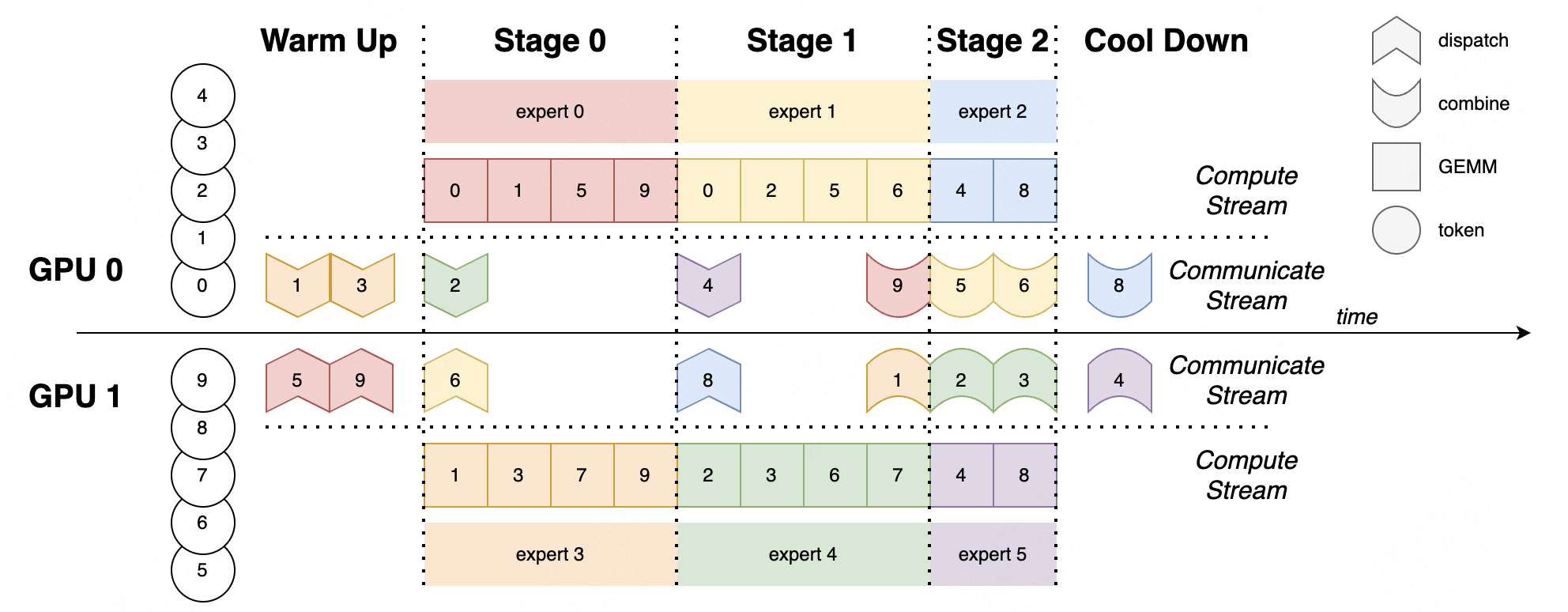}  
        \caption{Expert Pipeline Scheduler. Expert number=6, topk=2, DP=2, EP=2. The outputs of Attention Blocks on GPU0 and GPU1 are token [0,1,2,3,4] and token [5,6,7,8,9] respectively. Expert0 takes token [0,1,5,9], Expert1 takes token [0,2,5,6], and so on.}
        \label{fig:eps_overview}
    \end{subfigure}
    \hfill
    \begin{subfigure}[b]{0.45\textwidth}
        \centering
        \includegraphics[width=1\linewidth]{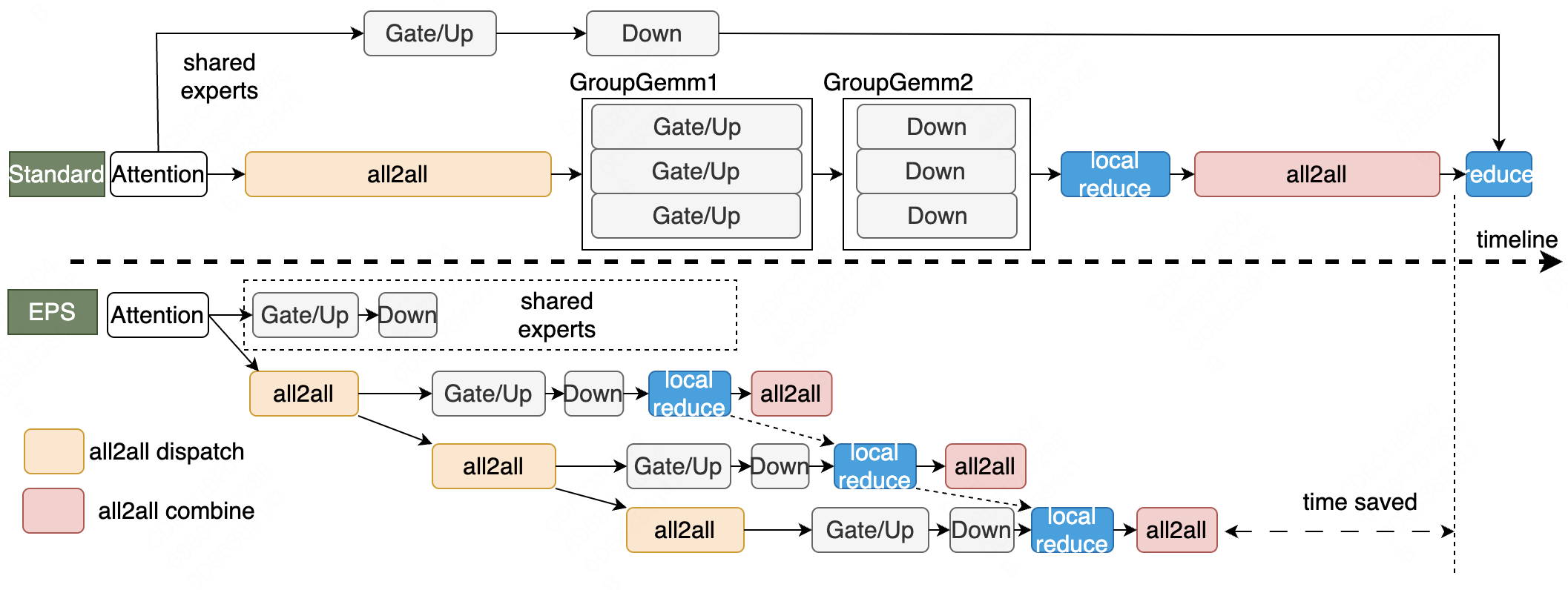}  
        \caption{Overlapping of communication and computation. \textit{LocalReduce} is memory-bound operation. When the computation for all experts of a token is finished, the next round of all2all communication for this token can be initiated. }
        \label{fig:comp_overlap_comm}
    \end{subfigure}
    \hfill

    \caption{Illustration of Expert Pipeline Scheduler.}
    \label{fig:three_figures}
\end{figure*}

Figure~\ref{fig:three_figures} shows how EPS-MoE works on DP + EP. It demonstrates the expert-level pipeline with communication and computation overlapping. EPS-MoE consists of three core modules: Parallel Strategy, Expert Pipeline Scheduler (EPS) and Computation and Communication Overlapping.

\textbf{Parallel Strategy} \quad The parallel strategy determines the communication mode we adopt, and the communication mode dictates how we apply the overlap strategy. The choice of parallel strategy should be based on memory footprint, memory access, and changes in computation. Due to the unique nature of distributed expert computation, the MoE model introduces new changes to the parallel mode.

\textbf{Expert Pipeline Scheduler} \quad To apply the pipeline strategy, we need a method to partition the MoE input tensor and MoE weights matrix. The traditional TP partitioning method is one option. In EPS-MoE, we have designed another method to partition weights according to experts, based on the characteristics of MoE models. We will see that this partitioning method is necessary for applying more efficient GEMM computations and pipeline strategies.

\textbf{Computation and Communication Overlapping} \quad We split the input tensor by rows and only transmit the tokens required by the current experts to the corresponding device each time. By this method, we pipeline the computation and communication in parallel. We will discuss the design of overlapping strategy and its overheads in this section also.

\subsection{Parallel Strategy}

The MoE model mainly consists of two parts: Attention and MoE. From this point, the inference time of an MoE model can be described as below:
\[T_{LLM} = (T_{Attention} + T_{MoE} + T_{Comm}) \times L  \]

Attention can use TP (Tensor Parallelism) or DP (Data Parallelism), while MoE can adopt TP or EP (Expert Parallelism). Therefore, we have a lot of parallel strategies for MoE inference. We will discuss Attention TP + MoE TP, Attention DP + MoE EP and Attention TP + MoE EP in detail. The notations we will use are listed as below.

\begin{itemize}
    \item $F(\cdot)$ : Computation FLOPs of a certain block.
    \item $W(\cdot)$ : Memory I/O data volume of a certain block.
    \item $V_{S}(x)$: Communication volume of strategy $S$ for data volume $x$ on a device.\footnote{We only consider the output volume from a device.} $V_O$ is communication overhead.
    \item $M_{in}$,$M_{out}$: Input/output activation volumes per device; $M_{KVCache}$ : Total KVCache volume (0 during prefill); $M_{weight}$: Parameter volume. 
    \item $k$: Number of experts selected by a token in MoE model. 
    \item $D$: Device number of TP, EP or DP.
    \item $FP$, $BW$, $NW$: Computational FLOPS, memory bandwidth, and NVLink bandwidth between devices.
    \item $P$: Activation size for batched requests.
\end{itemize}

For different parallel strategies, the computation throughput of a single device is the same. The core impacts of different parallel strategies are: 1. the computing resources for a single token, considering the parallel mechanism of modern gpus; 2. the memory footprint on different devices, as some parallel strategies cause redundant storage of parameters across different devices; 3. the amount of memory access on different devices. So, the computational throughput of Attention or MoE depends on the maximum memory access volume and computation flops and the communication time.

\[T_{Attn} = max\{\frac{F(Attn)}{D \times FP}, \frac{W(Attn)}{D \times BW}\} \]

\[T_{MoE} = max\{\frac{F(MoE)}{D \times FP}, \frac{W(MoE)}{D \times BW}\}\]

\[T_{Comm} = max\{\frac{V_S(\cdot)}{NW}, V_O\}\]
\noindent
We will focus on discussing $W(\cdot)$ and $V(\cdot)$.

\paragraph{TP + TP} The Attention part uses the TP mode, and the MoE part also uses the TP mode. 
 
\[W(Attn) = M_{in}  + M_{out} + \frac{ [M_{KVCache}]}{D} + \frac{M_{weight}}{D}\]
\[W(MoE) = M_{in} \cdot k + M_{out} \cdot k + \frac{M_{weight}}{D}\]

As we can see, if the top-k of MoE is large, the impact caused by the activation value I/O is significant. 

Since all tokens are present on each device and \textit{ncclAllReduce} communication is used, the communication volume of one device for TP+TP is:

\[V_{TP+TP}(P, D)=2\frac{P}{D}(D-1).\]

\paragraph{DP + EP} The problem with using TP for the Attention part is that when Attention applies the MLA (Multi-head Latent Attention) structure, extra communication needs to be introduced. Therefore, we use DP to accelerate the MLA Attention part and EP for MoE.

\[W(Attn) = \frac{M_{in}  + M_{out} + [M_{KVCache}]}{D} + M_{weight}\]
\[W(MoE) = \frac{M_{in} \cdot k}{D} + \frac{M_{out} \cdot k}{D} + \frac{M_{weight}}{D}\]

For DP+EP, the communication time mainly depends on how the experts are deployed. If all experts selected by a token are on the same device, the communication volume is optimal. However, if all experts selected by a token are distributed across different devices, there will inevitably be an amplification in communication. So, DeepSeekV2 has designed a device-limited routing mechanism to restrict MoE-related communication costs~\cite{deepseekv2}. Let $g$ be the device number one token will be routed to, $g$ is bounded by $\min(k, D)$, i.e. $g \le \min(k, D)$. Therefore, the total data volume of communication of one device for DP+TP is listed below:

\[ \frac{P}{D^2}(D -1) \le V_{DP + EP}(P/D,D) \le g \frac{P}{D^2}(D - 1).\]

Therefore, we have 
\[V_{DP +EP}(P/D, D) \le V_{TP +TP}(P, D).\]

\paragraph{TP + EP} DP will result in deduped parameters storage between different devices, so for models whose attention is MHA (Multi Head Attention), GQA (Grouped Query Attention) or MQA (Multi Query Attention), it's better to use TP for attention part.

\[W(Attn) = M_{in}  + M_{out} + \frac{ [M_{KVCache}]}{D} + \frac{M_{weight}}{D}\]
\[W(MoE) = \frac{M_{in} \cdot k}{D} + \frac{M_{out} \cdot k}{D} + \frac{M_{weight}}{D}\]

As for communication, normally, we use \textit{ncclAllReduce} for TP communication. However, in order to apply expert pipeline, we decompose \textit{ncclAllReduce} into two steps to make better communication effeciency. For dispatch stage, we use \textit{ncclReduceScatter} + \textit{all2all} instead of \textit{ncclAllReduce}. For combine stage, we use \textit{all2all} + \textit{ncclAllGather} instead of \textit{ncclAllReduce}.This will make communication volume much less for MoE models. For dispatch stage, we have
\[V_{TP+EP}(P, D)=\frac{P}{D}(D-1) + V_{DP + EP}(P/D, D).\]

We could easily find that the communication volume of combine stage is the same as dispatch stage. By splitting the \textit{ncclAllReduce} operator in this way, we can parallelize the \textit{all2all} operator and GEMM using the expert pipeline strategy. Subsequently, we can also apply \textit{ncclReduceScatter} and \textit{ncclAllGather} with Flux\cite{chang2024flux} to integrate with the preceding and following GEMM operations.

\paragraph{Summary}
Different parallel patterns affect the amount of data accessed in memory and the communication data volume, but do not impact the time taken for computation. 
For memory accessed data volume of Attention block, which one is the best depends on the relationship  between attention weight and attention activation.
For memory accessed data volume of MoE block, we have
\[W_{DP + EP}(MoE) = W_{TP +EP}(MoE) < W_{TP+TP}(MoE).\]
For communication data volume, we have
\[ V_{DP+EP} < V_{TP+EP} < V_{TP+TP}.\]

DP does not change the overall computational throughput, but it leads to three issues: 1. Increased computation time for individual tokens; 2. Redundant storage of parameters; 3. Load balancing problems. For issue 1 and 2, we will trade them for improved throughput. For balancing problems, we will use chunked prefill to solve this problem. For EP balance strategy, methods such as balance loss or token drop~\cite{deepseekv2}\cite{gshard}\cite{m6-t} have already been studied, which we will skip here.

\subsection{Expert Pipeline Scheduler}

\begin{figure}[htbp]
\centering
\includegraphics[width=1\linewidth]{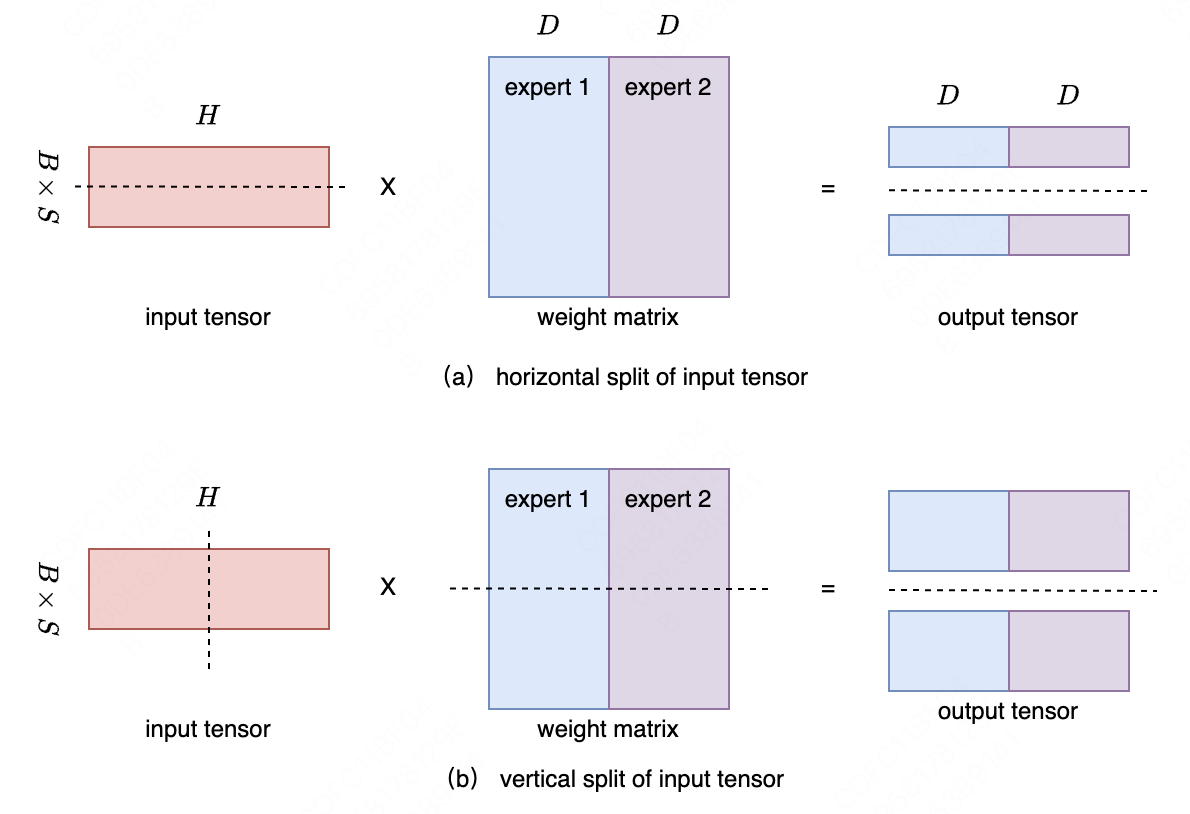}
\caption{Two different ways to split input tensor, horizontal split and vertical split. (a) Horizontal split. Split rows of input tensor. Weights split by experts. (b) Vertical split. Split cols of input tensor. Weights split by columns.}
\label{fig:two_way_to_split_input_tensor}
\end{figure}

There are two data partitioning methods that can achieve pipeline parallelism for computation and communication in MoE models. As shown in Figure~\ref{fig:two_way_to_split_input_tensor}, horizontal split organizes the input data by row partitioning on the input tensor, while vertical split organizes the input data by column partitioning on the input tensor. The traditional horizontal splitting method leads to repeated memory I/O of the parameter matrix. When applying horizontal splitting, we utilized the sparsity characteristics of MoE parameters and split the parameters according to experts, which can effectively avoid repeated memory I/O of the parameter matrix.

Based on this, let's consider the memory I/O efficiency differences between these two splitting methods. Let the number of samples be \( m \), the size of a single input and output activation be \( P_0 \) and \( P_1 \), the size of each expert parameters be \( W \), the total number of experts be \( E \), and the number of pipelines be \( N \). The total I/O data volume \( V \) is given by:

\[ V(a) = m \cdot P_0 + E \cdot W + m \cdot P_1 \]
\[ V(b) = m \cdot P_0 + E \cdot W + N \cdot m \cdot P_1 \]

Method (a) results in much less I/O data volume compared to method (b).
Let the computational workload of a single expert be \( C \), and the computational throughput of \textit{GroupGemm} with group size \( G \) be \( R(\text{FLOPS} | G) \). The computational time difference between the two methods is:

\[ T(a) = \frac{C \cdot E/N}{R(\text{FLOPS} | E/N)} \cdot N = \frac{E \cdot C}{R(\text{FLOPS} | E/N)} \]
\[ T(b) = \frac{C \cdot E/N}{R(\text{FLOPS} | E)} \cdot N = \frac{E \cdot C}{R(\text{FLOPS} | E)} \]

The final difference is reflected in the computational efficiency of \textit{GroupGemm} with different group sizes. When \( E = N \), \textit{GroupGemm} can be replaced by \textit{DenseGemm} to achieve better performance for larger input $m$. We can see that the core advantages of using a horizontal split are a more efficient GEMM implementation and a lower memory I/O data volume. Therefore, the Expert Pipeline Scheduler implements pipeline scheduling by horizontally splitting the input tensor. Meanwhile, the weights are partitioned according to the experts, and each time only the tokens required by a specific group of experts are transmitted.
From the previous analysis data, we can see that the comparative advantage of  \textit{GroupGemm} and \textit{DenseGemm} varies under different computational loads. Based on this, we designed a load-aware adaptive scheduling strategy to dynamically select different efficient implementations based on the type of load. 
Through this parallel strategy, we can relatively easily establish parallelism of token-level transmission and computation, as shown in Figure~\ref{fig:eps_overview}. 

\subsection{Computation and Communication Overlapping}

Pipeline parallelism is a common strategy to improve resource utilization when a task requires different types of hardware resources.\cite{aibox} Generally, when multiple hardware resources work together, pipeline parallelism can be used to enhance overall throughput. In the context of LLM inference, some kernels are memory-bound, some are compute-bound, and some are communication-bound. However, the hardware resource usage for each type of kernel requires the support of SM resources. Since some kernels' execution may require both memory access and computation, while others may require both memory access and communication, no single type of kernel can utilize all types of hardware resources to their theoretical maximum. From the previous analysis, we can infer that by controlling the number of SMs occupied by some kernels, we can leave a portion of SMs for other kernels while keeping the kernel's execution time unchanged or changed not too much. This is a fundamental prerequisite for implementing pipeline parallelism. 
In EPS-MoE, we parallelized both \textit{GroupGemm} and \textit{DenseGemm} with the \textit{all2all} communication to improve hardware resource utilization as shown in Figure~\ref{fig:comp_overlap_comm}. In subsequent tests, we will show that if we control the number of SMs occupied by computation and communication, the pipeline can achieve even greater benefits.

\section{Experiments and Ablations}

\subsection{Experiments}

We tested expert pipeline on various kinds of MoE models. For MoE models with fine-grained experts, we choose DeepSeekV2 as an example. For MoE models with less experts, we selected Mixtral 8x7B and DBRX. For MoE models with an overlapping structure between computation and communication such as Snowflake Arctic, we also tested EPS-MoE on such a model to demonstrate why expert pipeline is efficient.

\begin{table}[htbp]
\caption{Prefill Throughput Gains of DeepSeekV2.}
\label{tab:prefill_benefit}
\centering
\begin{threeparttable}
\begin{tabular}{l|rrrr}
    \toprule
    PN\textbackslash seq & 1K & 2K & 4K & 8K \\
    \midrule
    PN=1 & +11.83\% & +11.02\% & +6.77\% & +5.13\% \\
    PN=5 & +21.81\% & +20.53\% & +16.89\% & +12.75\% \\
    PN=20 & +21.27\% & +18.05\% & +10.00\% & +11.60\% \\
    \bottomrule
\end{tabular}
\begin{tablenotes}
    \item[1] PN=1 indicates that pipeline is not used, but \textit{GroupedGemm} will be switched to \textit{DenseGemm}.
\end{tablenotes}
\end{threeparttable}
\end{table}

\paragraph{Experiments on DeepSeekV2}
We tested the effect of expert pipeline on the DeepSeekV2 on 8xH800-80GB SXM. In this test, we used the same methods and environment as DeepSeekV2 to test the maximum prefill throughput. The DeepSeekV2 model is a fine-grained MoE model with outstanding performance in both effectiveness and efficiency. As \cite{moe_scaling_laws} has pointed out, the more granularity an MoE model has, the better loss it will achieve. And our method will take advantages of the granularity of MoE model. In Table~\ref{tab:prefill_benefit}, we present the prefill throughput benefits of expert pipeline on DeepSeekV2 for different Pipeline Numbers (PN). It can be seen from the table that the prefill throughput benefit can reach up to 21\%. Given that DeepSeekV2 has already provided a relatively high baseline (with a reported prefill throughput of 100,000 tokens/s), this improvement can be considered a significant result. Note that we have applied Attention DP + MoE EP on DeepSeekV2 as its attention is MLA.

\begin{table}[h]
\caption{TTFT Reduction of Mixtral8x7B and DBRX.}
    \centering
    \label{tab:mixtral}
    \begin{threeparttable}
    \begin{tabular}{c|c|rrrr}
        \hline
         Mixtral & \textbf{B\textbackslash S} & \textbf{2k} & \textbf{4k} & \textbf{8k} & \textbf{16k} \\ 
        \cline{1-6}
        \multirow{4}{*}{\textbf{base}} 
        & 16 & 0.46 & 0.94 & 1.94 & 4.22 \\
        & 32 & 0.92 & 1.86 & 3.88 & - \\
        & 64 & 1.83 & 3.71 & -& -\\
        & 128 & 3.65 & - & - & - \\

        \hline
        \multirow{4}{*}{\textbf{TP+EP}} 
        & 16 & -13.7\% & -9.5\% & -8.9\% & -8.3\% \\
        & 32 & -10.3\% & -9.4\% & -9.0\% & - \\
        & 64 & -9.5\% & -9.3\% & -& -\\
        & 128 & -9.6\% & - & - & - \\

        \hline
        \multirow{4}{*}{\textbf{PN=1}} 
        & 16 & -17.7\% & -18.9\% & -18.5\% & -15.7\% \\
        & 32 & -19.6\% & -19.2\% & -18.1\% & - \\
        & 64 & -19.7\% & -18.9\% & -& -\\
        & 128 & -19.5\% & - & - & - \\

         \hline
        \multirow{4}{*}{\textbf{PN=2}}  
        & 16 & -21.7\% & -22.7\% & -21.8\% & -17.7\% \\
        & 32 & -23.6\% & -23.0\% & -22.6\% & - \\
        & 64 & -23.4\% & -23.5\% & - & - \\
        & 128 & -24.3\% & - & - & - \\
         \hline
        DBRX & \textbf{B\textbackslash S} & \textbf{0.5k} & \textbf{1k} & \textbf{2k} & \textbf{4k} \\ 
        \hline
        
        \multirow{3}{*}{\textbf{base}} 
        & 16 & 0.22 & 0.42 & 0.84 & 1.68  \\
        & 32 & 0.42 & 0.83 & 1.66 & 3.35 \\
        & 64 & 0.84 & 1.67 & 3.32 &  - \\
        \cline{1-6}
        \multirow{3}{*}{\textbf{PN=1}}  
        & 16 & -19.8\% & -21.8\% & -24.4\% & -25.1\%  \\
        & 32 & -22.0\% & -24.2\% & -25.3\% & -25.7\%  \\
        & 64 & -24.1\% & -25.7\% & -26.1\% & - \\
        \cline{1-6}
        \multirow{3}{*}{\textbf{PN=2}}  
        & 16 & -22.7\% & -25.7\% & -28.6\% & -29.5\% \\
        & 32 & -25.3\% & -28.7\% & -30.0\% & -30.3\%  \\
        & 64 & -28.7\% & -30.3\% & -30.5\% & - \\
        \hline
    \end{tabular}
   \begin{tablenotes}
    \item[1]  Base data units are in seconds. 
    \item[2] B for batch size. S for sequence length.
    \end{tablenotes} 
\end{threeparttable}
\end{table}

\paragraph{Experiments on Mixtral8x7B and DBRX}

We deployed Mixtral8x7B on 4xH800-80GB SXM and DBRX on 8xH800-80GB SXM, using Attention TP + MoE EP from vLLM. We integrated EPS-MoE with vLLM. The baseline is the TP + TP implementation of vLLM and kernels have been tuned for the best performance. From Table~\ref{tab:mixtral}, we can see that EPS-MoE has at most 24.3\% prefill latency reduction on Mixtral8x7B. Compared to the performance of DeepSeekV2, the gains of EPS-MoE on Mixtral8x7B increase with the sequence length. This is because the DeepSeekV2 model is relatively large, and as the sequence length increases, the maximum batchsize we can use decreases. Expert pipeline, on the other hand, benefits more from larger batchsizes. But for seqlen larger than 16K, the Attention compute time will be too much more to be ignored, so the benefit of EPS-MoE will decrease. 
By comparing the TP+EP results, which use the same GEMM implementation with baseline, with the base results in the test data of Mixtral8x7B, it can be seen that switching from the TP+TP parallelism mode of vLLM to the TP+EP parallelism mode of EPS-MoE, and replacing the AllReduce communication primitive with ReduceScatter and All2All, resulted in 8\% \textasciitilde 13\% reduction in TTFT. Compare the data of PN=2 and PN=1, it can be found that the switch of GEMM and concurrency mode from TP+TP to TP+EP contributed to at most 19.5\% reduction while pipelined overlapping contributed to 4.8\% reduction. Besides, compared to Mixtral8x7B, the DBRX model has more experts and a larger topk. EPS-MoE will benefit from having more experts or a larger topk. Therefore, EPS-MoE can achieve greater benefits on DBRX.

\paragraph{Experiments on Snowflake Arctic}

\begin{table}[h]
\caption{TTFT Reduction of Snowfake Arctic.}
    \centering
    \label{table:dbrx_and_arctic}
    \begin{threeparttable}
    \begin{tabular}{c|c|c|rr}
        \hline
         \textbf{model} & TTFT & \textbf{bs \textbackslash seq} & \textbf{2k} & \textbf{4k}  \\ 
        \cline{1-5}

        \multirow{6}{*}{Snowflake}
        & \multirow{2}{*}{\textbf{base}}  & 16 & 23.50 & 41.94 \\
        & & 32 & 40.84 & 82.62  \\

        \cline{2-5}
        & \multirow{2}{*}{\textbf{PN=4}}  & 16 & -0.26\% & -0.57\% \\
        & & 32 & -2.15\% & -6.24\%  \\

        \cline{2-5}
        & \multirow{2}{*}{\textbf{PN=8}}  & 16 & -1.26\% & -0.96\% \\
        & & 32 & -3.83\% & -6.64\%  \\

        \hline     
    \end{tabular}
   \begin{tablenotes}
    \item[1] Base data units are in milliseconds (ms).
    \item[2] Snowflake Arctic tested with 3 layers.
    \end{tablenotes} 
\end{threeparttable}
\end{table}

We also tested the application of the expert pipeline method in models with a parallel structure, such as Snowflake Arctic. The test platform used 8xH800-80G SXM. From Table~\ref{table:dbrx_and_arctic}, it can be seen that the expert pipeline method can achieve certain benefits for this parallel model structure given a specific batch size. The analysis found that these benefits are mainly due to switching the matrix operations method from \textit{GroupGemm} to \textit{DenseGemm}. The benefits of pipeline are relatively small.


\begin{table*}[htbp]
\centering
\begin{threeparttable}
\caption{Performance  of \textit{GroupGemm} and \textit{DenseGemm} with Different Pipeline Numbers.}
\label{tab:perf_gemm_overlap}
\begin{tabular}{lccccccccccc}
    \toprule
     ID & PN & GEMM & Overlapping & FP8 & SM & \multicolumn{2}{c}{m=3072} & \multicolumn{2}{c}{m=1024} & \multicolumn{2}{c}{m=256} \\
    \cmidrule(lr){7-8} \cmidrule(lr){9-10} \cmidrule(lr){11-12}
    & & & & & &  1536 & 15360 & 1536 & 15360 & 1536 & 15360 \\
    \midrule
0 & 1 & GroupGemm & N & N & 132 & 5.698 & 32.469 & 2.120 & 8.996 & 1.032 & 2.349 \\
1 & 1 & GroupGemm & N & Y & 132 & 4.330 & 30.980 & 1.512 & 8.516 & 1.513 & \textbf{2.203} \\
2 & 1 & DenseGemm & N & N & - & 5.620 &  21.247 & 3.889 & 7.497 & 0.868 & 2.708 \\
3 & 5 & GroupGemm & Y & N & 132 & 4.837 & 25.979 & 1.790 & 8.333 & 0.683 & 2.452 \\
4 & 5 & GroupGemm & Y & N & 116 & 4.248 & 26.270 & 1.568 & 8.808 & 0.636 & 2.461 \\
5 & 5 & GroupGemm & Y & Y & 132 & 3.728 & 24.833 & 1.418 & 7.996 & 0.607 & 2.358 \\
6 & 5 & GroupGemm & Y & Y & 116 & 3.277 & 25.928 & \textbf{1.299} & 8.697 & \textbf{0.597} & 2.398 \\
7 & 5 & DenseGemm & Y & N & - & 4.299 & 20.203 & 1.714 & 7.128 & 0.798 & 2.598 \\
8 & 5 & DenseGemm & Y & Y & - & \textbf{3.111} & \textbf{19.795} & 1.438 & 6.895 & 0.727 & 2.549 \\
9 & 20 & DenseGemm & Y & N & - & 4.855 & 20.599 & 2.133 & 7.043 & 1.016 & 2.655 \\
10 & 20 & DenseGemm & Y & Y & - & 3.465 & 20.012 & 1.715 & \textbf{6.893} & 0.977 & 2.619 \\
    \bottomrule
\end{tabular}
\begin{tablenotes}
    \item[1] FP8 means whether the communication datatype is FP8 or not. 
    \item[2] SM means the SM count of GEMM. We do not control the SM count of DenseGemm.
    \item[3] All data units are in milliseconds (ms).

\end{tablenotes}
\end{threeparttable}
\end{table*}

\subsection{Ablations}

We adopted the model structure of DeepSeekV2, using one layer consist of Gate, Up, Down GEMM as the test subject to analyze the following issues: 1. The benefits of switching from DenseGemm to GroupedGEMM; 2. The impact of FP8 communication; 3. The necessity of controlling SM; 4. The impact of different output dimensions; 5. The impact of the number of pipelines as is shown by PN. These test data are uniformly presented in Table~\ref{tab:perf_gemm_overlap}. In Table~\ref{tab:perf_gemm_overlap}, the input dim is 5120, the output dim is 1536 or 15360, the former is the same size as DeepSeekV2, the latter is three time of input dim as Llama-like models.\cite{touvron2023llama2openfoundation} We use 4xH800-80GB SXM to do the test.

\noindent
\textbf{1. Contributions of Overlapping and GEMM switching.}

Switching GEMM and overlapping both yield certain benefits, especially when the number of input tokens is large. From Table~\ref{tab:perf_gemm_overlap}, it can be seen that when the input size is relatively large, such as when $m=3072$ and GEMM size is [5120, 1536], if switching GEMM from \textit{GroupGemm} to \textit{DenseGemm} can bring a 12\% benefit (from 4.837 ms in data [ID=3] to 4.299 ms in data [ID=7]), while the benefit of overlapping is 16\% (from 5.698 ms in data [ID=0] to 4.837 ms in data [ID=3]). When the input size is small, the benefit from switching GEMM is smaller and may even be negative, with the main benefits coming from the pipeline strategy. It is important to note that when the input size is small and GEMM is memory-bound, EPS-MoE (w/o overlapping) has a slight advantage over the overlapping setting. This is because the communication overhead is much heavier as discussed before.

\noindent
\textbf{2. Does FP8 communication reduce the benefits of expert pipeline?}

In most cases, FP8 communication reduces the proportion of time spent on communication, thereby affecting the benefits of overlapping strategy. However, FP8 communication also faces precision issues due to quantization. By simultaneously applying FP8 communication and overlapping strategy, we can see that these two methods can further combine to generate greater benefits. From  Table~\ref{tab:perf_gemm_overlap}, it can be seen that FP8 communication and the expert pipeline strategy are not in conflict, and in most cases, using both FP8 communication and the expert pipeline strategy together can yield further benefits. For example, considering \(m=3072\) and GEMM size is [5120, 1536], comparing the data [ID=2] and data [ID=8], both of whose communication data type is set to be FP8, the latter still shows a 28.2\% improvement over the former.

\begin{table}[htbp]
\caption{Time Cost Under Different SM.}
\label{tab:control_sm}
\centering
\begin{threeparttable}
\begin{tabular}{ccccccc}
    \toprule
    \multicolumn{1}{c}{$m\backslash SM$} & 132 & 124 & 116 & 108 & 100 & 92 \\
    \midrule
    3072 & 4.597 & 4.403 & \textbf{3.894} & 3.967 & 4.031 & 4.081 \\
    1536 & 2.385 & 2.336 & \textbf{2.079} & 2.130 & 2.162 & 2.054 \\
    768 & 1.306 & 1.268 & \textbf{1.170} & 1.180 & 1.196 & 1.167 \\
    384 & 0.894 & 0.921 & \textbf{0.787} & \textbf{0.780} & 0.818 & 0.797 \\
    192 & 0.591 & 0.596 & \textbf{0.540} & 0.541 & 0.591 & 0.797 \\
    \bottomrule
\end{tabular}
\begin{tablenotes}
    \item[1] SM: the SM count of GEMM. $m$: the input count of GEMM per expert.
    \item[2] All data units are in milliseconds (ms).
\end{tablenotes}
\end{threeparttable}
\end{table}

\noindent
\textbf{3. The necessity of controlling the number of SM.}

It is important to control the number of SM used by computation kernels when we apply expert pipeline, as we can see from both Table~\ref{tab:perf_gemm_overlap} and  Table~\ref{tab:control_sm}. In Table~\ref{tab:control_sm}, we tested one MoE layer consist of Gate, Up, Down GEMM with expert pipeline scheduling on 4xH800-80GB SXM. Controlling SM is mainly about managing the number of SMs occupied by computation when communication and computation overlap. The goal is to allocate more SMs to communication operators without affecting computation efficiency. The number of SMs allocated is crucial for the time consumption of both computation and communication. When the GEMM operator occupies all SMs, the communication operator cannot be scheduled, slowing down its execution. Additionally, the scheduling of the communication operator impacts the GEMM computation, increasing its execution time as well. In such cases, the benefits of overlapping are reduced. However, by controlling the number of SMs occupied by the GEMM operator, communication and computation operators can be executed in parallel without interfering with each other, achieving perfect overlap.
Besides, in Table~\ref{tab:control_sm}, we have set the SM numbers of the communication kernel to be 16, so the best SM numbers for computation in these tests are all 116.

\noindent
\textbf{4. When should we enable expert pipeline?}

Expert pipeline does not work in all scenarios. If the number of tokens is small, expert pipeline will fail. Generally speaking, as long as the time spent on communication and computation after division is less than that before division, Expert pipeline will be beneficial. However, since the division is not linearly variable in terms of computation and communication in some areas, it is relatively difficult to quantify and analyze. We can calculate a relatively loose lower bound.

Lets define the communication time function is $S(m ; k,n)$, the MoE GroupGemm time function is $G(m,E;k,n)$, where $m$ is input token number, $k, n$ is input dim and output dim, $E$ is expert number, $N$ is pipeline number and $1\le N \le E, E\mid N$. We can find that expert pipeline will work as long as for a certain $\epsilon_0, \epsilon_1 \in R$, we have:

\[\left\{ m \mid \frac{N \cdot S\left(\frac{m}{N} ; k,n\right)}{S(m ; k,n)} \leq 1 + \epsilon_0 \text{ and } \frac{N \cdot G\left(\frac{m}{N}, \frac{E}{N} ; k, n\right)}{G(m, E ; k, n)} \leq 1 + \epsilon_1 \right\}.\]

\begin{figure}
    \centering
    \includegraphics[width=1\linewidth]{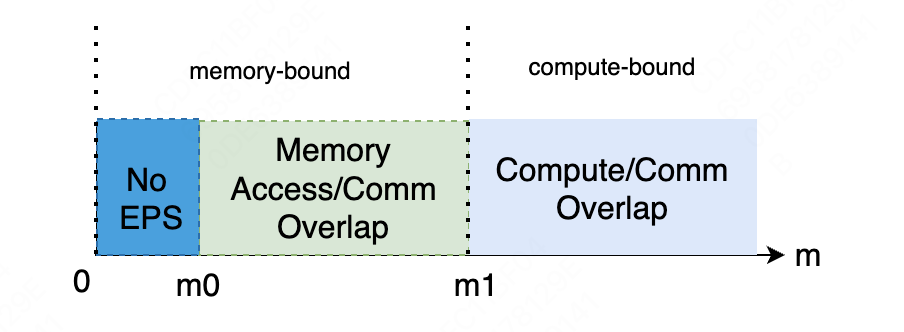}
    \caption{The range where EPS-MoE works.}
    \label{fig:benifit_range}
\end{figure}

In the test of DeepSeekV2, we can find $m\ge 1700$ to satify the condition for $\epsilon_0 = 0.32, \epsilon_1=0.03$. From Table~\ref{tab:baseline_and_pipeline}, we can see that 1700 is a critical point with some marginal benefits. Note that the tests here are merely illustrative; since no kernel tuning has been done, the benefits will not exhibit linear variation with changes in $m$. The range of $m$ where EPS-MoE can beat original inference can be listed in Figure~\ref{fig:benifit_range}. When $m < m_0$, EPS-MoE would not work as the $\epsilon_0$ for communication operation is too big. When $m_0 < m < m_1$, the GEMM operation is memory-bound, EPS-MoE benefits from memory access and communication overlap. When $m > m_1$, EPS-MoE benefits from computation and communication overlap.

\begin{table}[h]
    \centering
    \caption{Comparison of Baseline and Pipeline.}
    \label{tab:baseline_and_pipeline}
    \begin{threeparttable}
    \begin{tabular}{cccccr}
        \toprule
        \textbf{m} & $\epsilon_0$ & $\epsilon_1$&\textbf{Base} & \textbf{Pipeline} & \textbf{Delta} \\
        \midrule
        1700 & 0.32 & 0.02 & 0.869 & 0.801 (PN=2) & -7.8\% \\
        1800 & 0.2 & 0 & 0.995 & 0.801 (PN=2) & -9.4\% \\
        1900 & 0.24 & 0 & 1.032 & 0.968 (PN=2) & -6.2\% \\
        2000 & 0.07 & 0 & 1.054 & 0.985 (PN=2) & -6.5\% \\
        2200 & 0.02 & 0.03 & 1.124 & 1.085 (PN=2) & -3.5\% \\
        \bottomrule
    \end{tabular}
    \begin{tablenotes}
        \item[1] Base and Pipeline are tested forward time of one layer on 8xH800 SXM. Both are in milliseconds(ms). Delta equals (Pipeline - Base)/Base.
        \item[2] $\epsilon_0$ indicates the communication time varies a lot when token is relatively small.
    \end{tablenotes}
    \end{threeparttable}
\end{table}

\noindent
\textbf{5. How to find the best PN number?}

\begin{figure*}[htbp]
    \centering
    \begin{subfigure}[b]{0.32\textwidth}
        \includegraphics[width=\linewidth]{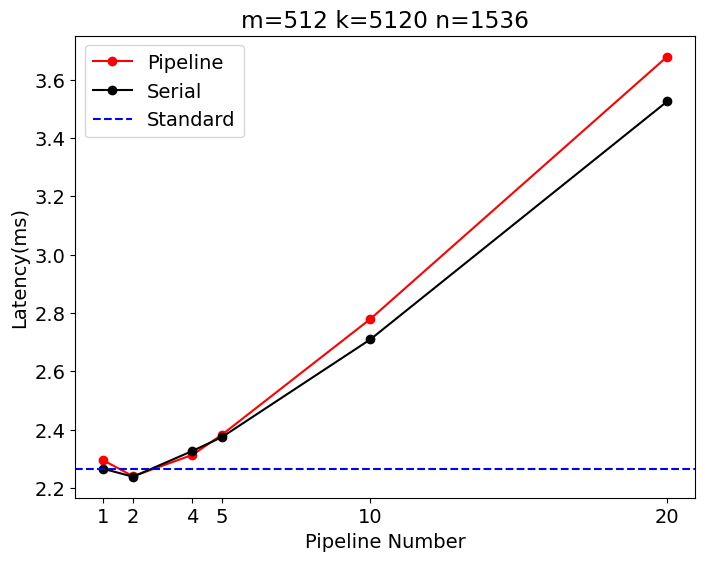}
        \caption{GEMM size is [5120, 1536]}
        \label{fig:pn128}
    \end{subfigure}
    \hfill
    \begin{subfigure}[b]{0.32\textwidth}
        \includegraphics[width=\linewidth]{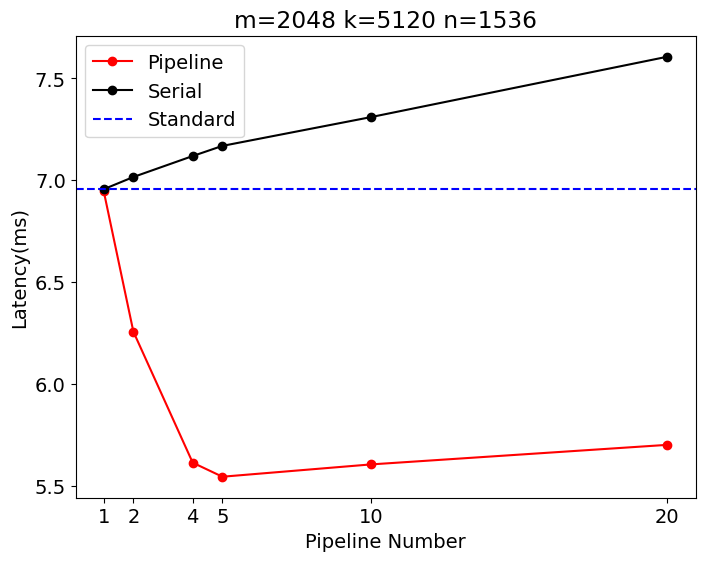}
        \caption{GEMM size is [5120, 1536]}
        \label{fig:pn3072}
    \end{subfigure}
    \hfill
    \begin{subfigure}[b]{0.32\textwidth}
        \includegraphics[width=\linewidth]{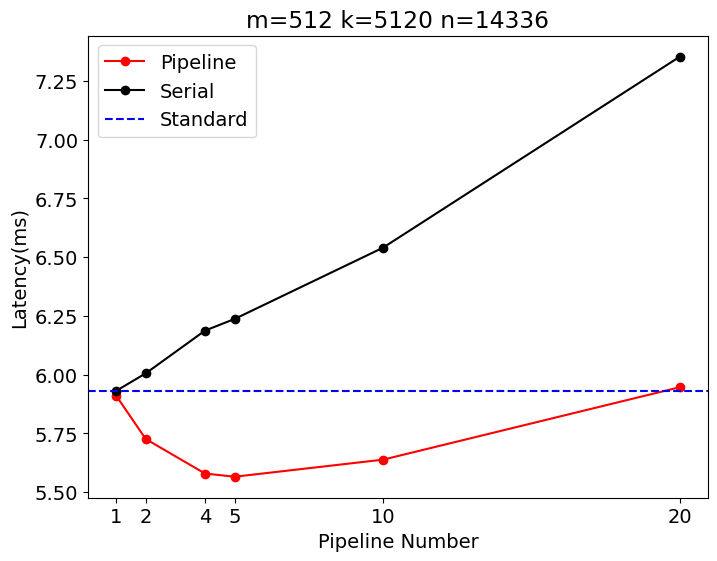}
        \caption{GEMM size is [5120, 14336]}
        \label{fig:pn256}
    \end{subfigure}
    \hfill
    \caption{Latency of an MoE kernel consisted of Gate/Up/Down GEMM with varying pipeline numbers. Tested on 4x H800 SXM NVLink GPUs. Standard means baseline with \textit{DenseGemm}. Here, we ignore the benefits of switching from \textit{GroupGemm} to \textit{DenseGemm}. Serial mode will split communication and computation into chunks and processes communication and computation chunks serially, showing partitioning overhead. For (a) and (b)input dim is 5120, output dim is 1536 (same as DeepSeekV2). For (c), input dim is 5120, output dim is 14336 (almost 3x like in Llama model). }
    \label{fig:pipeline_number}
\end{figure*}

The pipeline number is crucial for the acceleration of EPS-MoE, as shown in Figure~\ref{fig:pipeline_number}. Different pipeline numbers can lead to completely different results. From Figure~\ref{fig:pipeline_number}, we found that:
\begin{itemize}
    \item Given a certain inference task load and model computation cost, there exists an optimal Pipeline Number that maximizes the benefits of the pipeline parallel.
    \item When the number of tokens is relatively small, it is unlikely to achieve benefits regardless of how the pipeline is divided. This has been discussed in former section.
\end{itemize}
Different levels of granularity in chunks of input tensor will lead to changes in the ratio of communication and computation time. At the same time, due to some fixed overhead in partitioning, when the number of tokens is small, the benefits of the pipeline are decreased by the overhead. When the overhead is rarely small as in Figure~\ref{fig:pn3072}, the benefit of expert pipeline will be much more significant. Otherwise, EPS-MoE works inconspicuously as the overhead is too big. The pipeline number is influenced by a combination of factors such as hardware, kernel implementation, model parameters, and inference task load. During the optimization process for the DeepSeekV2 model, we first identified the optimal pipeline number through test data analysis and then applied it to our scenario.

We can establish a method through analysis to find the optimal pipeline number. We analyze the optimal pipeline number setting based on the time consumption of a single layer. Let $\theta$ represent the model parameters, $N$ represent the pipeline number, and $E$ represent the number of experts on a single device, such that \(1 \le N \le E\) and \(E \mid N\). Let function $L(\theta; N)$ represent the theoretical maximum benefit after overlapping computation and communication, and $R(N)$
 represent the overhead from pipeline. We determined from our test results that the overhead is linearly related to the pipeline number, which means $R(N) \approx kN + b$. Let $T_{comm}$ denote the total communication time before splitting the pipeline, and $T_{comp}$ denote the total computation time of the MoE layer before splitting. Finding the optimal pipeline number can be formulated by solving the following equation:

 \[\underset{1 \le N \le E}{argmax} \quad L(\theta;N) - R(N)\]
 
 where,

\begin{align}
L(\theta; N) &= \min\left\{ \frac{T_{\text{comm}}}{N}, \frac{T_{\text{comp}}}{N} \right\} \cdot (N-1) . \notag
\end{align}
 Let \(C=\min\{T_{\text{comm}} ,T_{\text{comp}}\}\), and $G$ for total latency reduction for overlapping. 

 \begin{align}
G &= C - b - \left(\frac{C}{N} + kN \right) \notag \\
&\le C - b - 2\sqrt{k \cdot C} \notag
\end{align}

 The maximum latency benefit is achieved if and only if \(N=\sqrt{\frac{C}{k}}\). The existence of overhead implies that there is an optimal pipeline number to achieve the maximum effect of the overlapping strategy. Otherwise, the more pipelines there are, the better the overlapping effect should be.

\section{Conclusion}
We designed different parallel schemes for different kinds of MoE models and implemented the Expert Pipeline Scheduler to make computation and communication overlapped for MoE models. Besides the overlapping strategy, switching from GroupedGemm to DenseGemm at a certain token number is also an important feature to improve throughput. We tested our method on different kinds of MoE models like DeepSeekV2, Mixtral8x7B and Snowflake Arctic. Experimental results demonstrate the effectiveness of the proposed solution and provide a new approach for further optimization of MoE inference.

\bibliographystyle{plain}
\bibliography{\jobname}

\end{document}